\documentclass{article}

\usepackage[utf8]{inputenc} 
\usepackage[T1]{fontenc}    
\usepackage{hyperref}       
\usepackage{url}            
\usepackage{booktabs}       
\usepackage{amsfonts}       
\usepackage{nicefrac}       
\usepackage{microtype}      
\usepackage[margin=1.2in]{geometry}

\usepackage{amsmath,amssymb,amsthm,graphicx,url}
\usepackage{algorithm}
\usepackage{algorithmic}

\usepackage{subfigure}

\usepackage{multirow}

\title{Implementation of Stochastic Quasi-Newton's Method in PyTorch}

\author{
  Yingkai Li \thanks{Authors contributed equally}\\
  Department of Computer Science\\
  Northwestern University\\
  \texttt{yingkai.li@u.northwestern.edu}  \\
  \and
  Huidong Liu \footnotemark[1] \\
  Department of Computer Science\\
  Stony Brook University \\
  \texttt{huidliu@cs.stonybrook.edu}
}

\begin{document}

\maketitle

\begin{abstract}
In this paper, we implement the Stochastic Damped LBFGS (SdLBFGS) for stochastic non-convex optimization. We make two important modifications to the original SdLBFGS algorithm. First, by initializing the Hessian at each step using an identity matrix, the algorithm converges better than original algorithm. Second, by performing direction normalization we could gain stable optimization procedure without line search. Experiments on minimizing a 2D non-convex function shows that our improved algorithm converges better than original algorithm, and experiments on the CIFAR10 and MNIST datasets show that our improved algorithm works stably and gives comparable or even better testing accuracies than first order optimizers SGD, Adagrad, and second order optimizers LBFGS in PyTorch.
\end{abstract}

\section{Introduction}
 
Current popular optimization methods for Deep Neural Networks (DNNs) are mainly first order with heuristics, e.g., SGD \cite{bottou2010large}, Adagrad \cite{duchi2011adaptive}, Adam \cite{kingma2014adam}, RMSProp \cite{tieleman2012lecture}, AdaDelta \cite{zeiler2012adadelta}, etc. Since DNNs are typically non-convex, it is difficult to apply the second order methods in optimizing DNNs due to the fact that the Hessian matrix might not be positive semi-definite (PSD), although second order methods converges much faster than first order methods. 

There has been attempts in applying the second order method Limited memory BFGS (LBFGS) to optimizing DNNs in the famous deep learning framework PyTorch. However, when optimizing DNNs, the Hessian of LBFGS may become negative semi-definite and thus the objective loss will increase to a very large value. Moreover, the optimization process of LBFGS is highly sensitive to the learning rate. If the learning rate is too large, the training process is unstable. Otherwise, the objective function does not decrease. 
Recently, Wang et al \cite{wang2017stochastic} propose a second order algorithm, Stochastic damped LBFGS (SdLBFGS), to optimize non-convex functions. The Hessian in each step of SdLBFGS is guaranteed to be PSD. However, the convergence is not fully proved and we observed non-convergence and crashes in experiments. 

Based on the original SdLBFGS algorithm, we modified two important steps in the algorithm: 
\begin{itemize}
\item[1] The Hessian in each step is initialized by the identity matrix.
\item[2] Normalize the direction computed in each step using $l_2$ normalization.
\end{itemize}
We implemented the original algorithm and our improved algorithm in PyTorch \footnote{Code is available at https://github.com/harryliew/SdLBFGS}. We denoted the original algorithm as SdLBFGS0 and our improved algorithm as SdLBFGS. 
By initialize the Hessian at each step using the identity matrix, SdLBFGS converges better than SdLBFGS0. By normalizing the direction, SdLBFGS is much more stable than SdLBFGS0 without line search. 

In the following sections, we will first briefly introduce the SdLBFGS0 algorithm. Then, we will show our improvements of the SdLBFGS algorithm. 
Moreover, we will do experiments on a simple 2D non-convex function, CIFAR10 \cite{krizhevsky2009learning} and MINST \cite{lecun1998gradient} datasets. We will compare our implementation with SGD, Adagrad, LBFGS in PyTorch. Finally we will conclude our paper. 

\section{The Original SdLBFGS0 Algorithm}
Since SdBFGS0 is based on LBFGS with minor modifications, we shall first introduce LBFGS.
\subsection{Limited memory BFGS (LBFGS)}
Denote by $d$ the dimension of the optimum solution. Since BFGS needs to store the Hessian in each step, the space complexity of BFGS in each step is $O(d^2)$, which is very expensive when optimizing DNNs. The Limited memory BFGS (LBFGS) is proposed to resolve this issue. Instead of storing the Hessian, LBFGS computes the approximation of inverse of Hessian using previously calculated variables $p$, $s_j$ and $y_j$ by 
\begin{eqnarray}
\label{eq: inv_H_lbfgs}
 H_{k,i} &=& (I - \rho_j s_j y^T_j) H_{k, i-1} (I - \rho_j y_j s^T_j) + \rho_j s_j s^T_j, 
 j = k-(c-i+1), \forall 1\leq i \leq c.      
 \end{eqnarray}
where $c = \min\{k,p\}$, $s_j = x_j - x_{j-1}$ and $y_j = g_j - g_{j-1}$, where $g_j$ is the sub-gradient in step $j$. Therefore, the space complexity in each step is reduced from $O(d^2)$ to $O(pd)$.

\subsection{Original Stochastic dampled LBFGS (SdLBFGS0)}
SdLBFGS0 adopts the idea of SdLBFGS, but substitutes the $y_j$ in Eq. (\ref{eq: inv_H_lbfgs}) to $\bar{y}_j$, where $\bar{y}_j$ is computed as 
\begin{eqnarray}
\label{eq: y_bar}
 \bar{y}_{k-1} = \theta_{k-1} y_{k-1} + (1-\theta_{k-1}) H_{k,0}^{-1}s_{k-1} 
 \end{eqnarray}
where 
$$
 \theta_{k-1} = \left\{ \begin{array}{ll}
 \frac{0.75 s_{k-1}^T H_{k,0}^{-1}s_{k-1} }{s_{k-1}^T H_{k,0}^{-1}s_{k-1} - s_{k-1}^T y_{k-1}} &\mbox{ if $s_{k-1}^T y_{k-1} < 0.25 s_{k-1}^T H_{k,0}^{-1}s_{k-1}  $} \\
  1 &\mbox{ otherwise}
       \end{array} \right.
$$
In SdLBFGS0, 
\begin{eqnarray}
\label{eq: H_0}
H_{k,0} = \gamma_{k} ^ {-1} I ~~~~~\mbox{where}~~\gamma_{k} = \max \left \{ \frac{ y_{k-1}^T y_{k-1}}{ s_{k-1}^T y_{k-1} }, \delta \right \} > 0
 \end{eqnarray}
where $\delta > 0 $ is a given constant. 
By substituting $y_j$ by $\bar{y}_j$, the Hessian computed by SdLBFGS0 is guaranteed to be PSD. 

The step size in \cite{wang2017stochastic} is set to be 
\begin{equation*}
\alpha_k = \frac{\underline{\kappa}}{L\bar{\kappa}^2} k^{-\beta},
\end{equation*}
where $\beta \in (0.5, 1)$, $L$ is the Lipschitz constant 
and $\underline{\kappa}, \bar{\kappa}$ are the bounds on the Hessian. 
The authors claim that the number of iterations required for convergence is $O(\epsilon^{-\frac{1}{1-\beta}})$. 
The idea for computing the approximation of Hessian is using the stochastic damped L-BFGS (SdLBFGS) \cite{liu1989limited}. 

\section{Our Improvement of the SdLBFGS Algorithm}
In this section, we will introduce our two important improvements of the SdLBFGS0, 1) Initializing the Hessian with an identity matrix, and 2) Normalizing the direction in each step. The algorithm of SdLBFGS is presented in Algorithm \ref{alg:SdLBFGS}. Note that in Procedure 3.1 of \cite{wang2017stochastic}, the subscripts in line 9-12 are out of range. 

\begin{algorithm}[htbp]
  \caption{Step computation using SdLBFGS}
 \label{alg:SdLBFGS}
  \begin{algorithmic}[1]
\REQUIRE Let $x_k$ be a current iterate. 
		Given the stochastic gradient $g_{k-1}$ at iterate $x_{k-1}$, 
		the random variable $\xi_{k-1}$, the batch size $m_{k-1}$, $s_j$, $\bar{y}_j$ 
		and $\rho_j$, $j = k-p, \dots, k-2$, and $u_0 = g_k$. 

\ENSURE $H_kg_k = v_p$.

\STATE Set $s_{k-1} = x_k - x_{k-1}$ and $y_{k-1} = g_k - g_{k-1}$. 
\STATE Set $H_{k,0} = I$
\STATE Calculate $\bar{y}_{k-1}$ through Eq. (\ref{eq: y_bar}) and $\rho_{k-1} = (s_{k-1}^T \bar{y}_{k-1})^{-1}$

\STATE Set $c = \min\{p,k-1\}$
\FOR{$i = 0, \dots, c-1$} 
	\STATE Calculate $\mu_i = \rho_{k-i-1} u_i^T s_{k-i-1}$. 
	\STATE Calculate $u_{i+1} = u_i - \mu_i \bar{y}_{k-i-1}$. 
\ENDFOR 
\STATE Set $v_0 = u_p$
\FOR{$i = 0, \dots, c-1$} 
	\STATE Calculate $\nu_i = \rho_{k-c+i} v_i^T \bar{y}_{k-c+i}$. 
	\STATE Calculate $v_{i+1} = v_i + (\mu_{c-i-1} - \nu_i) s_{k-c+i}$. 
\ENDFOR 
\STATE $v_p = v_p / ||v_p||_2$
\end{algorithmic}
\end{algorithm}

\subsection{Initialize the Hessian with an Identity Matrix}
SdLBFGS0 proposed in \cite{wang2017stochastic} cannot converge to a local minimum in practice. 
A crucial difference between SdLBFGS and the original LBFGS is that the original LBFGS method applies backtracking line search to guarantee  convergence,  
while SdLBFGS0 uses the diminishing step size instead of line search. 
We would prefer using diminishing step size in SdLBFGS0, because when training DNNs, 
the computation of line search is highly expensive. 
Without line search, an immediate consequence of SdLBFGS0 is that it cannot converge to a local minimum in practice. The main reason is that, in SdLBFGS0, if the Hessian is large, then at some point the algorithm will output a direction with a small norm.
With diminishing step size, 
the movement towards that direction is small. 
This will also cause the the norm of direction in the next step to be small. 
Therefore, with diminishing step size, this algorithm will slowly converge to some point which is not a local minimum. Therefore, $\delta$ in Eq.~(\ref{eq: H_0}) needs to be chosen very carefully.
To rectify this situation, we will set the initial approximation of the Hessian matrix $H_{k,0}$ in each step $k$ to be the identity matrix $I$. 
In this way, the norm of the direction will not converge to zero 
even with diminishing step size, and
the objective will still gradually converge to a local minimum. This step is line 2 in Algorithm \ref{alg:SdLBFGS}.

\subsection{Direction Normalization}

Another problem of SdLBFGS0 is that the norm of the direction might be too large in some cases.
In this case, without line search, the movement towards that direction may be too large, especially when the algorithm runs in the first few steps where the step size is still large. Thus, this will cause the algorithm to be very unstable.
Hence, the objective fuction may become increasingly larger and gradually approach infinity. 
To address this issue, 
we normalize the direction in each step using the~$l_2$ norm, so that in each step, the algorithm will not move too far away from the current point. This makes the algorithm stable. This step is stated in line 14 of Algorithm \ref{alg:SdLBFGS}.

\section{Experiments}
In the experiments, we analyze the performance of our algorithm on a simple 2D non-convex function, the CIFAR10 \cite{krizhevsky2009learning} and the MNIST \cite{lecun1998gradient} datasets. Denote the implementation of the original algorithm in \cite{wang2017stochastic} as SdLBFGS0, and our modification of SdLBFGS0 as SdLBFGS. We compare our algorithm with the build-in optimizers SGD, Adagrad, LBFGS in PyTorch, and we implement both SdLBFGS0 and SdLBFGS in PyTorch. For SdLBFGS0 and SdLBFGS, we set the step size to be $1/\sqrt{k}$, where $k$ is the number of iterations. The memory sizes of LBFGS, SdLBFGS0 and SdLBFGS are all set to be~100 for fair comparison. The batch sizes for all the those methods are set to be 64 on CIFAR10 and MNIST datasets. 

\subsection{Optimization of A simple 2D Non-Convex Function}

\begin{figure}[htbp]
  \begin{center}
  \includegraphics[width=60mm]{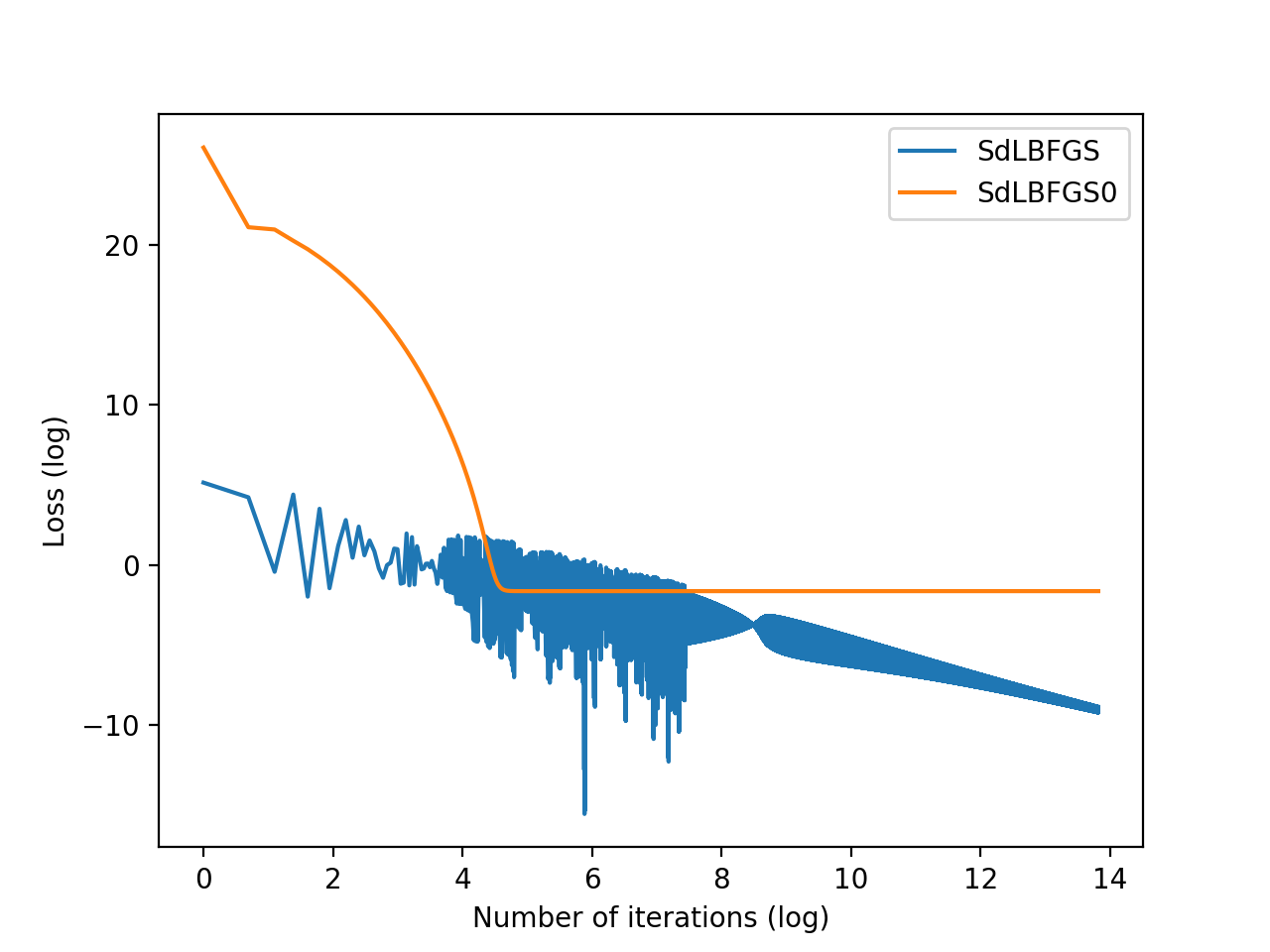}
  \caption{SdLBFGS0 and SdLBFGS on a simple 2D non-convex function.}
  \label{fig: simple_case}
  \end{center}
  \vspace{-.5cm}
 \end{figure}

In this experiment, we compare SdLBFGS0 and SdLBFGS on a simple 2D non-convex function expressed as follows:
 $$
	f(x, y ) = 100(x^2-y)^2 + (x-1)^2
 $$
For both methods, we start from $(-1.2, 1)$ and perform 1 million iterations. We show the number of iterations (in natural log) vs. the objective function value (in natural log) in Figure \ref{fig: simple_case}. From this figure we can see that the the value of the objective function optimized by SdLBFGS0 is very high at the beginning, because the direction of SdLBFGS0 is not normalized, and the iteration point goes too far away from the optimum. In addition, after about 100 number of iterations, the value of the objective function no longer decreases although the diminishing step size is applied. This means that SdLBFGS0 does not converge. By contrast, the value of the objective function optimized by SdLBFGS is normal at the beginning, and it decreases as the number of iterations increases. This demonstrates that our modification of the original algorithm is effective. 
\subsection{Results on the CIFAR10 dataset}
\begin{figure}[t!]
\centering     
\subfigure[]{\label{fig:subfig:cifar10_sdlbfgs_all_acc}\includegraphics[width=60mm]{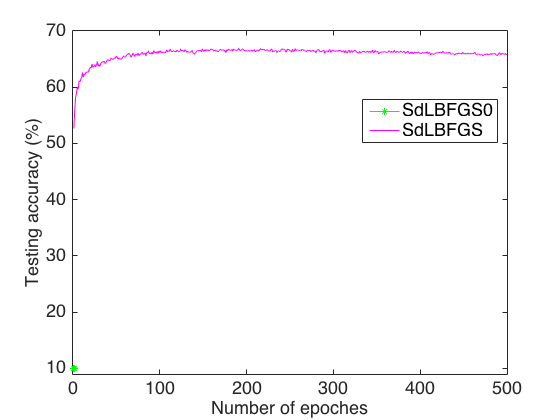}}
\subfigure[]{\label{fig:subfig:cifar10_sdlbfgs_all_loss}\includegraphics[width=60mm]{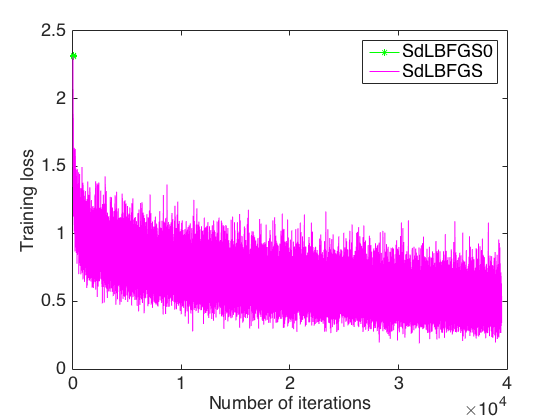}}
\caption{Comparison of SdLBFGS0 and SdLBFGS on the CIFAR10 dataset.}
\label{fig: cifar10_sdlbfgs}
\vspace{-.5cm}
\end{figure}

\begin{figure}
\centering     
\subfigure[]{\label{fig:subfig:cifar10_sgd_acc}\includegraphics[width=60mm]{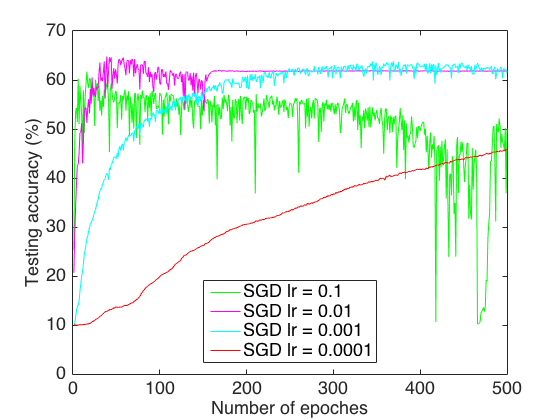}}
\subfigure[]{\label{fig:subfig:cifar10_adagrad_acc}\includegraphics[width=60mm]{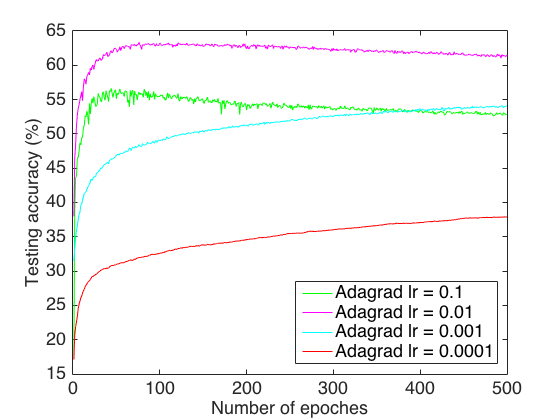}}
\subfigure[]{\label{fig:subfig:cifar10_lbfgs_acc}\includegraphics[width=60mm]{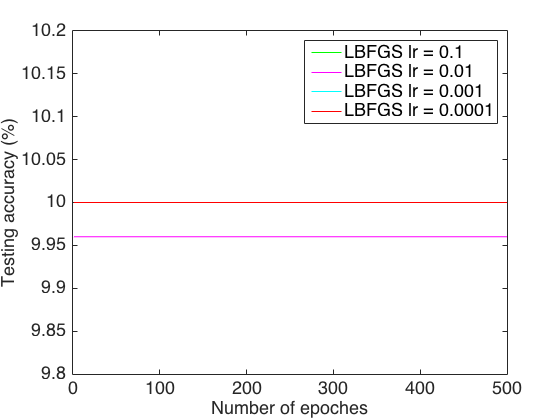}}
\subfigure[]{\label{fig:subfig:cifar10_sdlbfgs_acc}\includegraphics[width=60mm]{CIFAR10_SdLBFGS_all_acc.png}}
\caption{Accuracies of SGD, Adagrad, LBFGS, SdLBFGS0 and SdLBFGS under different learning rates on CIFAR10 dataset. Note that in (c) LBFGS lr = 0.1 overlaps lr = 0.01, and lr = 0.001 overlaps lr = 0.0001.}
\label{fig: cifar10_all4methods_acc}
\vspace{-0.5cm}
\end{figure}

\begin{figure}
\vspace{-.5cm}
\centering     
\subfigure[]{\label{fig:subfig:cifar10_sgd_loss}\includegraphics[width=60mm]{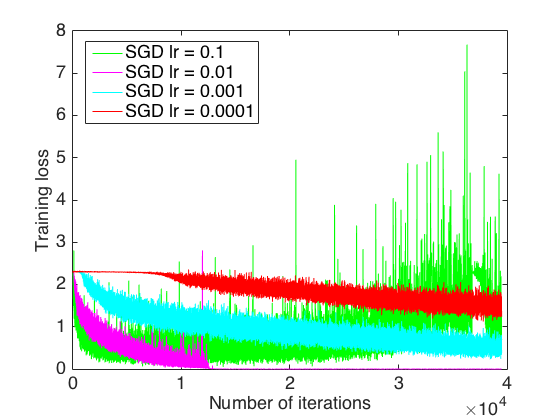}}
\subfigure[]{\label{fig:subfig:cifar10_adagrad_loss}\includegraphics[width=60mm]{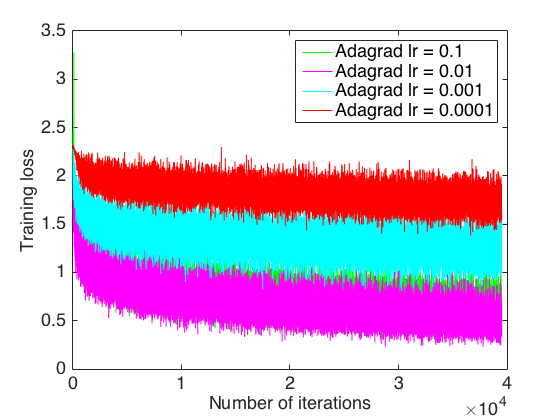}}
\subfigure[]{\label{fig:subfig:cifar10_lbfgs_loss}\includegraphics[width=60mm]{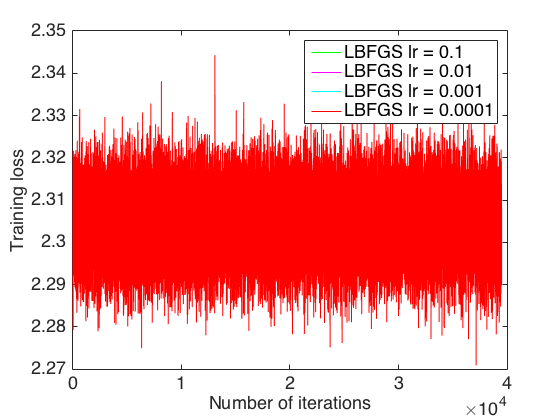}}
\subfigure[]{\label{fig:subfig:cifar10_sdlbfgs_loss}\includegraphics[width=60mm]{CIFAR10_SdLBFGS_all_loss.png}}
\caption{Losses of SGD, Adagrad, LBFGS, SdLBFGS0 and SdLBFGS under different learning rates on the CIFAR10 dataset.}
\label{fig: cifar10_all4methods_loss}
\vspace{-.5cm}
\end{figure}

We apply different optimizers to do classification on the CIFAR10 dataset. We use the convolution neural network provided online \footnote{https://pytorch.org/tutorials/beginner/blitz/cifar10\_tutorial.html} for all the optimizers. For SGD, Adagrad and LBFGS we choose their learning rates (lr) from [1e-4, 1e-3, 1e-2, 1e-1]. Figure \ref{fig: cifar10_all4methods_acc} (a), (b) and (c) show the accuracies of different methods v.s. number of iterations under different learning rates. In Figure \ref{fig: cifar10_sdlbfgs} (d), SdLBFGS0 only works at the beginning. After a few epochs, it outputs nan and stops working, whereas our improved SdLBFGS works normally, i.e., the accuracy increases and remains stable. 
Figure \ref{fig: cifar10_all4methods_loss} (a), (b) and (c) show the losses of different methods v.s. number of iterations under different learning rates. From Figure \ref{fig: cifar10_all4methods_loss}(c), we can see that the build-in optimizer LBFGS in PyTorch does not converge. The loss of SdLBFGS0 in Figure \ref{fig: cifar10_all4methods_loss}(d)
is higher than that of SdLBFGS and SdLBFGS0 crashes after a few number of iterations. This shows that SdLBFGS0 is not stable. 

\begin{figure}
\centering     
\subfigure[]{\label{fig:subfig:cifar10_comp_all4methods_acc}\includegraphics[width=60mm]{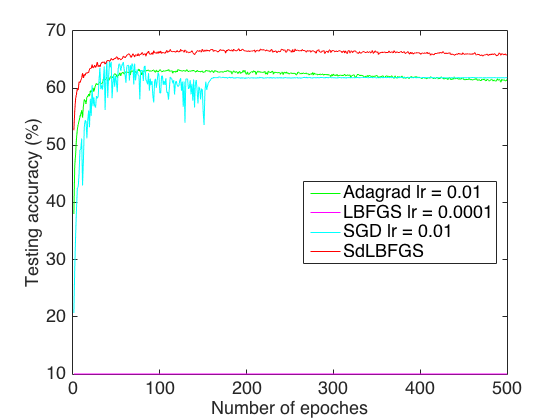}}
\subfigure[]{\label{fig:subfig:cifar10_comp_all4methods_loss}\includegraphics[width=60mm]{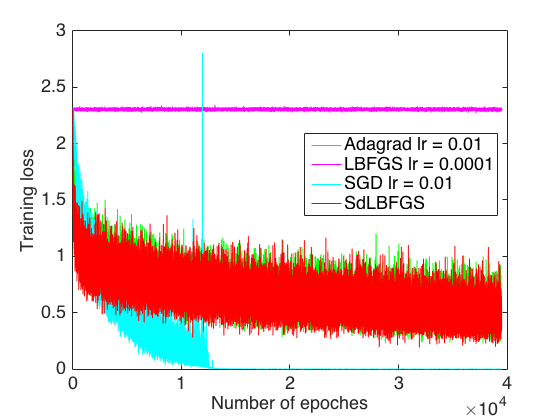}}
\caption{Comparison of different methods under their best learning rates on the CIFAR10 dataset.}
\label{fig: cifar10_comp_all4methods}
\vspace{-.5cm}
\end{figure}

In Figure \ref{fig: cifar10_all4methods_loss}, we choose the best learning rate for different methods and compare them together in accuracies (Figure \ref{fig:subfig:cifar10_comp_all4methods_acc}), and in losses (Figure \ref{fig:subfig:cifar10_comp_all4methods_loss}). From Figure \ref{fig:subfig:cifar10_comp_all4methods_acc} we can see that the testing accuracy of our implementation SdLBFGS increases faster than all the other methods under their best learning rates. Besides, SdLBFGS achieves the highest final testing accuracy of approximately 66\% compared to all the other methods. SGD and Adagrad achieve approximately similar testing accuracy of approximately 62\%. LBFGS gets the lowest accuracy, since the loss does not decrease according to Figure \ref{fig:subfig:cifar10_comp_all4methods_loss}. The losses of SdLBFGS and Adagrad are similar in Figure \ref{fig:subfig:cifar10_comp_all4methods_loss}. Although the loss of SGD decreases to nearly 0 in Figure \ref{fig:subfig:cifar10_comp_all4methods_loss}, it's testing accuracy is less than SdLBFGS.

\subsection{Results on the MNIST dataset}

\begin{figure}
\centering     
\subfigure[]{\label{fig:subfig:mnist_sdlbfgs_all_acc}\includegraphics[width=60mm]{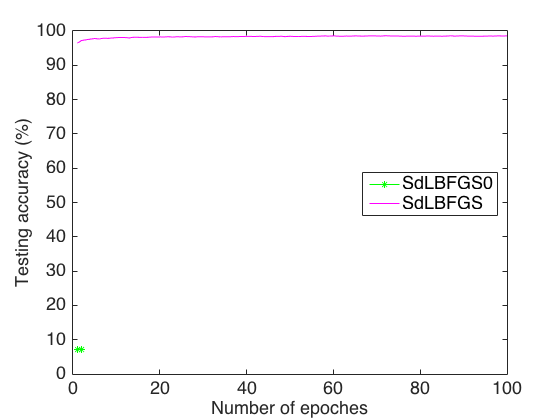}}
\subfigure[]{\label{fig:subfig:mnist_sdlbfgs_all_loss}\includegraphics[width=60mm]{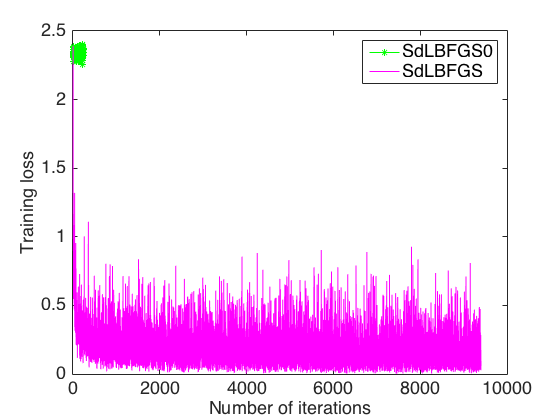}}
\caption{Comparison of SdLBFGS0 and SdLBFGS on the MNIST dataset.}
\label{fig: mnist_sdlbfgs}
\vspace{-.5cm}
\end{figure}

\begin{figure}
\centering     
\subfigure[]{\label{fig:subfig:mnist_sgd_acc}\includegraphics[width=60mm]{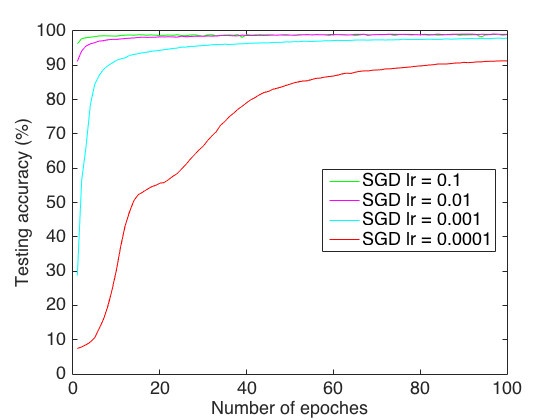}}
\subfigure[]{\label{fig:subfig:mnist_adagrad_acc}\includegraphics[width=60mm]{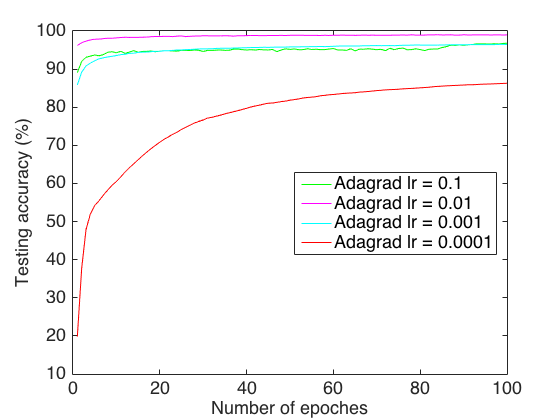}}
\subfigure[]{\label{fig:subfig:mnist_lbfgs_acc}\includegraphics[width=60mm]{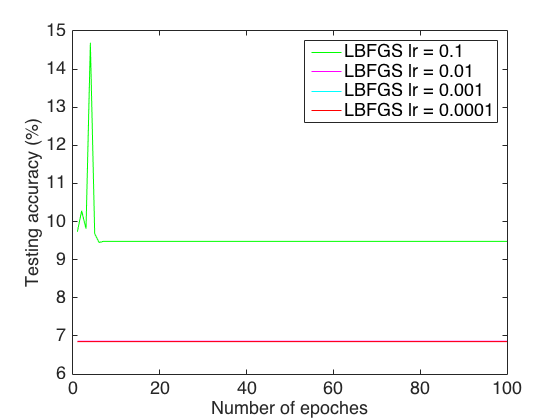}}
\subfigure[]{\label{fig:subfig:mnist_sdlbfgs_acc}\includegraphics[width=60mm]{MNIST_SdLBFGS_all_acc.png}}
\caption{Accuracies of SGD, Adagrad, LBFGS, SdLBFGS0 and SdLBFGS under different learning rates on the MNIST dataset. Note that in (c) LBFGS lr = 0.01 overlaps lr = 0.001 and lr = 0.0001.}
\label{fig: mnist_all4methods_acc}
\vspace{-.5cm}
\end{figure}

\begin{figure}
\centering     
\subfigure[]{\label{fig:subfig:mnist_sgd_loss}\includegraphics[width=60mm]{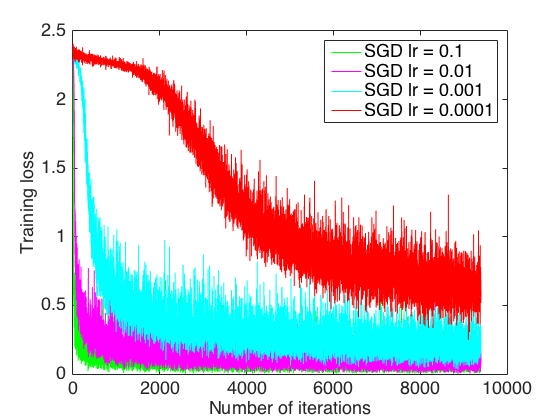}}
\subfigure[]{\label{fig:subfig:mnist_adagrad_loss}\includegraphics[width=60mm]{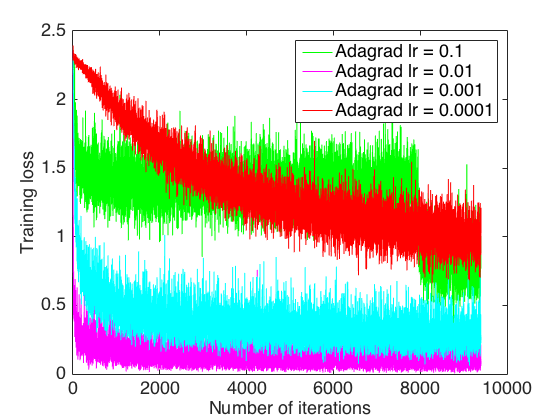}}
\subfigure[]{\label{fig:subfig:mnist_lbfgs_loss}\includegraphics[width=60mm]{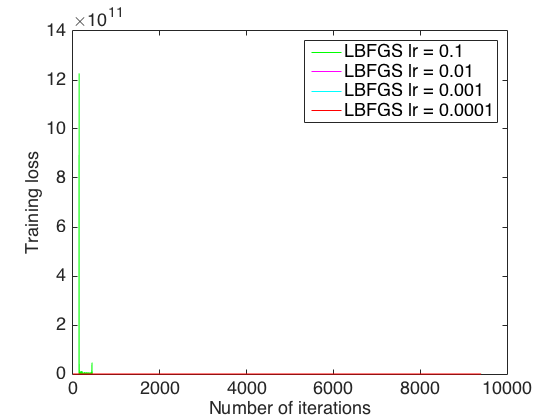}}
\subfigure[]{\label{fig:subfig:mnist_sdlbfgs_loss}\includegraphics[width=60mm]{MNIST_SdLBFGS_all_loss.png}}
\caption{Losses of SGD, Adagrad, LBFGS, SdLBFGS0 and SdLBFGS under different learning rates on the MNIST dataset.}
\label{fig: mnist_all4methods_loss}
\vspace{-.6cm}
\end{figure}

Similar to the experiments on the MNIST dataset, we apply different optimizers to do classification on the MNIST dataset. We use the convolution neural network provided on line \footnote{https://github.com/pytorch/examples/tree/master/mnist} for all the optimizers. 
The learning rates of SGD, Adagrad and LBFGS are chosen from [1e-4, 1e-3, 1e-2, 1e-1]. 
Figure~\ref{fig: mnist_all4methods_acc}~(a), (b) and (c) show the accuracies of different methods v.s. number of iterations under different learning rates. 
From Figure \ref{fig: mnist_sdlbfgs}(d) we can see that SdLBFGS0 works at the beginning, but after a few epochs, it crashes, whereas SdLBFGS works stably. Figure \ref{fig: mnist_all4methods_loss} (a), (b) and (c) show the losses of different methods v.s. number of iterations under different learning rates. Figure \ref{fig: mnist_all4methods_loss}(c) shows that LBFGS is very sensible to learning rates. Figure \ref{fig: mnist_all4methods_loss}(d) shows that SdLBFGS0 stops working at the beginning of the training and the loss does not decrease. 

\begin{figure}
\centering     
\subfigure[]{\label{fig:subfig:mnist_comp_all4methods_acc}\includegraphics[width=60mm]{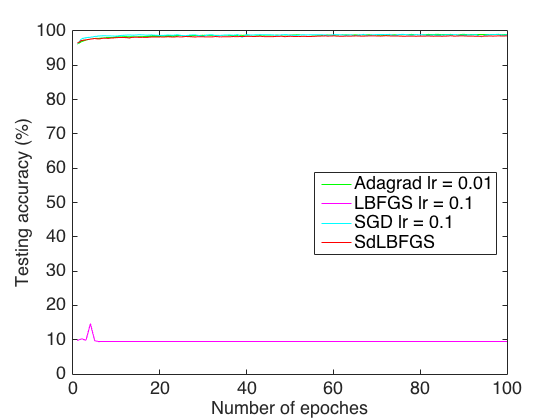}}
\subfigure[]{\label{fig:subfig:mnist_comp_all4methods_loss}\includegraphics[width=60mm]{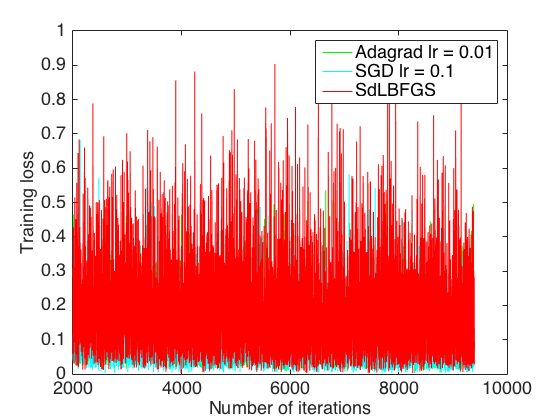}}
\caption{Comparison of different methods under their best learning rates on the MNIST dataset.}
\label{fig: mnist_comp_all4methods}
\end{figure}

In Figure \ref{fig: mnist_comp_all4methods}, SGD, Adagrad, LBFGS, SdLBFGS are compared under their best learning rates tuned in Figures \ref{fig: mnist_all4methods_acc} and \ref{fig: mnist_all4methods_loss}. From Figure \ref{fig:subfig:mnist_comp_all4methods_acc} we can see that the testing accuracies of SGD, Adagrad and SdLBFGS are comparable, which is approximately 98\%, whereas LBFGS performs poorly at about~10\%. The losses plotted in Figure \ref{fig:subfig:mnist_comp_all4methods_acc} show that SGD, LBFGS and SdLBFGS give comparable losses.

\section{Conclusion}
This paper presents the implementation of the Stochastic damped LBFGS (SdLBFGS) method in PyTorch for non-convex stochastic optimization problem. By initializing the Hessian in each step using the identity matrix, the algorithm converges better than the original algorithm. By performing direction normalization, we could obtain stable optimization procedure without line search. Experiments on minimizing a 2D non-convex function shows that our improved algorithm converges better than original algorithm, and experiments on the CIFAR10 and MNIST datasets show that out implementation works stably and gives comparable or even better testing accuracies than widely-used optimizers SGD, Adagrad, and the second order optimizer LBFGS in PyTorch.

\section*{Acknowledgement}
The authors would thank Professor Francesco Orabona for insightful discussions and helpful suggestions.

\medskip
\small 
\bibliographystyle{abbrv}
\bibliography{ref}

%
%

\end{document}